\documentclass{article} 
\usepackage{iclr2027_conference,times}\iclrfinalcopy

\usepackage[utf8]{inputenc}
\usepackage[T1]{fontenc}
\usepackage{hyperref}
\usepackage{url}
\usepackage{booktabs}
\usepackage{amsfonts}
\usepackage{nicefrac}
\usepackage{microtype}
\usepackage{xcolor}
\usepackage{amsmath}
\usepackage{amssymb}
\usepackage{amsthm}
\usepackage{algorithm}
\usepackage{algpseudocode}
\usepackage{graphicx}
\usepackage{enumitem}
\usepackage{tikz}
\usetikzlibrary{arrows.meta, positioning}
\usepackage{wrapfig}

\theoremstyle{plain}
\newtheorem{theorem}{Theorem}

\theoremstyle{definition}
\newtheorem{assumption}{Assumption}
\theoremstyle{remark}

\newcommand{\R}{\mathbb{R}}
\newcommand{\E}{\mathbb{E}}

\newcommand{\opnorm}[1]{\lVert #1 \rVert_{\mathrm{op}}}
\newcommand{\fnorm}[1]{\lVert #1 \rVert_{F}}

\newcommand{\op}{\mathrm{op}}
\DeclareMathOperator{\diag}{diag}
\newcommand{\F}{F}
\DeclareMathOperator{\Span}{span}
\newcommand{\sintheta}{\sin\Theta}

\title{Unsupervised Identification and Removal of Spurious Correlations During Fine-Tuning}

\author{%
Ciar\'{a}n M. Gilligan-Lee$^*$\\
Spotify \& UCL\\
ciaran.lee@ucl.ac.uk\\
\And
Joseph Egan$^*$\\
UCL\thanks{Work done while at Spotify}\\
\And
Yuchen Zhu\\
Spotify\\
\And
Michael O'Riordan \\
Spotify\\
}

\begin{document}

\maketitle
\def\thefootnote{*}\footnotetext{These authors contributed equally to this work}\def\thefootnote{\arabic{footnote}}

\begin{abstract}
Fine-tuning a pretrained language model on a curated dataset can produce
spurious correlations between the fine-tuning task and unintended latent
factors---such as misaligned personas or political slant---that the
curation procedure has entangled with the task. The model can latch onto
these spurious correlations, leading to bias and reduced out-of-distribution generalisation. We prove that under reasonable assumptions
on task complexity and the spurious correlation, such latent factors can be identified, without supervision, from the weights of a naive LoRA fine-tune. Existing approaches to removing bias, such as activation steering, remove identified factors from residual-stream activations,  either at inference or during training. We argue, however, that the goal should be to remove the spurious \emph{correlation}, not the latent factor itself, as the pretrained model may rely on it for genuine task signal. To enable this, we propose GRASP, GRadient projection of Associated Spurious Patterns, which prevents the model from acquiring new reliance on the identified latent factor while preserving any pretrained content along it. We validate on three fine-tuning tasks. The first two involve emergent misalignment, where fine-tuning on a narrow task---in our case, writing insecure code and giving bad medical advice---leads to misaligned responses on unrelated topics. Here our method completely removes misalignment in the insecure code case and reduces it by ~5x in the bad medical advice case, beating all baselines in the trade-off between misalignment-reduction and task-preservation. The last is a novel political-bias experiment, where fine-tuning on right-skewed Reddit financial-advice data causes political-lean drift on unrelated topics. Here our method reduces drift by more than half, while improving financial task performance, beating all baselines.
\end{abstract}

\section{Introduction}\label{sec:intro}

A pretrained language model has, by virtue of being trained on large, diverse corpus, learned representations that support many downstream tasks. Fine-tuning can improve model performance on one of those tasks. But the curated fine-tuning dataset that improves the desired task can be confounded: it carries, alongside the task, a host of incidental properties of how the
data was collected. Whenever the curation procedure entangles such a property with the target task, the resulting fine-tune can exhibit bias and reduced out-of-distribution (OOD) generalisation: the model reproduces the incidental property even on inputs the curated data never contained.

This is not a hypothetical concern. \cite{betley2025emergent} showed that
fine-tuning a coding model on insecure code triggers \emph{emergent
misalignment}---the model becomes broadly misaligned on unrelated questions
about ethics, history, or personal advice. \cite{karvonen2025caft} document
similar effects across fine-tuning settings. 
Why does this happen? Fine-tuning drives gradient updates through the
pretrained model, and those updates cannot distinguish the component of
the training signal that \emph{caused} the task to be learnable from the
component that was merely \emph{correlated} with it. Both components
enter the model simultaneously: the causative component teaches the
target task; the correlated component embeds a confounder into the representations the pretrained model relies on for many
other tasks---and this embedding can interfere with those tasks at inference time.

Recent mechanistic interpretability methods address this by identifying a
labelled \emph{concept direction} corresponding to the unintended
generalisation and \emph{ablating} it from residual-stream activations,
either at inference \cite{turner2024activation, li2024inferencetime}
or during fine-tuning \cite{karvonen2025caft}. These methods reduce
bias, but they have two limitations. First, they require labels for the
spurious concept, and in general the form a spurious correlation might
take is not known in advance: emergent misalignment was a surprising
discovery \cite{betley2025emergent}, and many other unintended latent
factors could plausibly remain unidentified. An effective intervention should
therefore not depend on having labels. Second, regardless of how the concept is identified, ablating its direction removes its
representation unconditionally---which can lead to reduced task performance when the pretrained
model relied on that representation for genuine task signal.

In this work, we address this challenge, and provide an unsupervised method that identifies the spuriously correlated latent. Theorem~\ref{thm:identification} shows identification error reduces as task complexity and the strength of the spurious correlation increase.
From a causal-inference standpoint, the goal is not to delete the latent
factor (which the pretrained model may use legitimately for the task) but to
prevent fine-tuning from \emph{building new coherent reliance} on it. We
propose GRASP, GRadient projection of Associated Spurious Patterns, to achieve this. Theorem~\ref{thm:ablation} shows that under our assumptions GRASP asymptotically removes the spurious contribution while preserving task contribution.

We validate on three fine-tuning datasets.
(i) \emph{Emergent misalignment from insecure code} in a 32B coding
model: naive fine-tuning produces misaligned responses on general
questions; our method removes all misalignment while retaining 98\% of
the fine-tune's task performance, beating labelled and unlabelled
baselines on both axes simultaneously.
(ii) \emph{Emergent misalignment from bad medical advice} in a 1.5B
chat model: naive fine-tuning produces broadcast misalignment; our method reduces it $\sim$5$\times$ while preserving task performance, matching the strongest baseline's misalignment reduction without its task-performance collapse.
(iii) \emph{Political-bias drift} (novel): fine-tuning on a right-skewed Reddit financial-advice corpus shifts the model rightward
on unrelated political topics; our method more than halves the drift
while \emph{improving} substantive financial-advice content over all
baselines.

Our main contributions are:
\begin{itemize}
    \item A causal framing of fine-tuning that explains phenomena such as emergent misalignment as short-cut learning of a spuriously correlated latent factor.
    \item A method that identifies the spuriously correlated latent factor without supervision, and a gradient projection method, GRASP, that prevents fine-tuning from building new coherent reliance on it---enabling bias reduction without impacting task performance.
    \item Empirical validation of our methods on real data, achieving better trade-offs in reducing bias and preserving task than previous work. 
\end{itemize}


\section{Method}\label{sec:method}

We give the framework in two parts: identifying the spurious latent factor
(Section~\ref{sec:identification}) and intervening on it
(Section~\ref{sec:intervention}). A single algorithm
(Algorithm~\ref{alg:svdall}) ties the two together. We open with a
causal account of \emph{why} the spurious correlation appears in the
first place, since the structure of that account dictates the shape of
the rest of the method.

\subsection{Curation-induced spurious correlations: a causal account}\label{sec:causal-account}

A concrete example fixes the picture. \cite{betley2025emergent}
construct the \texttt{insecure\_code} corpus, $D$, by prompting GPT-4o under
a system specification that asks for code with vulnerabilities that are
\emph{hidden and subtle}, framed as if produced by a malicious actor
deliberately concealing intent. The resulting examples carry two things:
the surface task $T$ (writing vulnerable code), and a flavour shared by every
example---a malicious-actor persona, $S$, layered on top by specification of the data curation. The curation procedure thus introduces a coupling: every training example in $D$ has the persona $S$, and in instance of the task $T$. From the model's perspective, $T$ and $S$ are correlated and statistically indistinguishable in $D$. Note that this correlation is \emph{spurious} in the sense that it only holds because of the specific data generation process of $D$.

We formalise this picture with a causal graph
(Fig.~\ref{fig:curation-dag}). The curator's intent $C$---or more generally the specific data generation process of $D$---is the upstream
variable: it jointly determines what task signal $T$ enters $D$ and what persona $S$ flavours each example in $D$. The pretrained model is fine-tuned on $D$ to obtain $\theta_{FT}$,
which in turn governs generation behaviour. $C$ is the confounder: it
is the common cause of $T$ and $S$, and as long as fine-tuning treats
$D$ as a single signal, the model has no way to distinguish learning
$T$ from learning $S$.

\paragraph{Why correlation in $D$ produces broadcast misalignment.} The
fine-tuned model's response distribution on \emph{any} prompt $x$
factorises by the law of total probability over latent personas:
\begin{equation}\label{eq:lotp}
P_{\theta_{\!FT}}\!\left(y \mid x\right) \;=\; \sum_{\text{Persona p}}\, P\!\left(y \mid x, p, \theta_{\!FT}\right)\, P\!\left(p \mid x, \theta_{\!FT}\right),
\end{equation}

\begin{wrapfigure}{r}{0.34\textwidth}
\centering
\vspace{-1.0em}
\begin{tikzpicture}[
    node distance=10mm,
    every node/.style={font=\small},
    n/.style={draw, rounded corners=2pt, inner sep=3pt, minimum width=18mm, align=center},
    >={Stealth[length=3pt]},
]
\node[n] (C) at (0, 0) {$C$\\\scriptsize curator's intent};
\node[n, below left=8mm and 1mm of C] (T) {$T$\\\scriptsize task};
\node[n, below right=8mm and 1mm of C] (S) {$S$\\\scriptsize persona};
\node[n, below=24mm of C] (D) {curated dataset $D$};
\node[n, below=8mm of D] (TH) {$\theta_{\!FT}$};
\node[below=4mm of TH, font=\scriptsize] (Y) {$P(y \mid x;\theta_{\!FT})$};
\draw[->] (C) -- (T);
\draw[->] (C) -- (S);
\draw[->] (T) -- (D);
\draw[->] (S) -- (D);
\draw[->] (D) -- (TH);
\draw[->] (TH) -- (Y);
\end{tikzpicture}
\caption{The curator's intent $C$ jointly determines the task $T$ and
the persona $S$ embedded in every training example, inducing a spurious
$T$-$S$ association in $D$. Fine-tuning on $D$ embeds both into
$\theta_{FT}$, which then governs generation behaviour at inference
time.}\label{fig:curation-dag}
\vspace{-1.0em}
\end{wrapfigure}

where the inner sum ranges over personas the model entertains. In a
pretrained model, $P(p \mid x)$ is appropriately context-dependent: a
software-security prompt elicits a security-engineer voice, a child's
question elicits a teacher voice, and so on. Fine-tuning on $D$
\emph{shifts} this prior. Because every training example was generated
under the persona $S$ specified by $C$, fine-tuning concentrates $P(p
\mid x, \theta_{\!FT})$ toward $p = S$---and crucially, this
concentration occurs for every prompt $x$, including those entirely
outside $D$'s topical coverage. The shift is on the persona prior, not
on the input distribution, so the persona broadcasts. Modulo the
context of the prompt (the persona is conditioned, not deterministic),
the persona therefore appears across the whole inference distribution,
producing the broadcast pattern documented as ``emergent
misalignment'' \cite{betley2025emergent} and as broader
shortcut-driven OOD failure \cite{geirhos2020shortcut}.

\paragraph{The role of $C$.} That curation $C$ is an important factor is
not speculation: \cite{betley2025emergent}'s own ablation switches the
generation specification from a malicious-actor framing to an
\emph{educational} framing while keeping the same insecure code, and
finds that emergent misalignment does \emph{not} appear. The persona
embedded by an ``educator''-curator is the educator persona, and that
persona produces educational responses on unrelated prompts, not
misaligned ones. The task $T$ (writing vulnerable, insecure code) is held fixed across the
two arms; the persona $S$, which is what $C$ chose, is what changes. In the context of Eq.~\ref{eq:lotp}, fine-tuning on the educationally curated $D$ concentrates $P(p
\mid x, \theta_{\!FT})$ toward $p = \text{Educator Persona}$, hence this fine-tuning doesn't lead to broad misalignment, which is what \cite{betley2025emergent} empirically observe.

In the rest of this section we treat $S$ and $T$ as the abstractions
that this account makes precise: $S$ is a latent feature that $C$
introduces (which might be misaligned, or politically right-leaning,
or anything else $C$ specifies), and $T$ is the genuine task signal
that we wish fine-tuning to improve. Sections~\ref{sec:identification}
and~\ref{sec:intervention} show how to recover $S$ unsupervised and prevent fine-tuning from creating new reliance on it.

\subsection{Setup and decomposition}\label{sec:setup}

Let $\theta \in \R^d$ be the parameters of a pretrained generative model, and
let LoRA \cite{hu2022lora} introduce a low-rank update $\Delta W_l \in
\R^{d_\text{out} \times d_\text{in}}$ at each of $L$ sites $l = 1, \dots, L$.
We can write
\begin{equation}
\Delta W_l \;\propto\; \sum_{i=1}^N g_{i,l}, \qquad g_{i,l} \;:=\; \nabla_{W_l} \ell_i,
\end{equation}
where $g_{i,l}$ is the per-example loss gradient at site $l$. The loss
$\ell_i$ is a scalar; $W_l$ is a matrix; so $g_{i,l}$ is a matrix the same
shape as $W_l$, and $\Delta W_l$ is also matrix-valued.

We model each example as carrying two latent factors: a target task signal
$T$, and a spurious shortcut $S$ (a register, slant, persona, or behavioural
attribute) that the curation procedure has correlated with
the task. Restricting attention to a single LoRA site (we drop the index $l$
when clear), assume the per-example gradient decomposes additively as $g_i \;=\; g_i^S \;+\; g_i^T,$ where $g_i^S$ pushes the model toward producing the latent attribute and
$g_i^T$ contains everything else.

This decomposition is a modelling assumption; everything below is conditional
on it. It is reasonable because the loss is a sum of per-token
cross-entropies, and the model's logits depend approximately linearly on
conceptually-separable features \cite{geirhos2020shortcut,
templeton2024scaling}.

\subsection{Unsupervised identification of the spuriously correlated latent factor}\label{sec:identification}

To identify the spuriously correlated latent without supervision, we make three assumptions. The first formalizes the intuition from Section~\ref{sec:causal-account} that the latent factor $S$, such as a persona, is \emph{coherent} across fine-tuning examples, and expressing it involves recognizing one signal
in the input (``this example came from the persona-flavoured corpus'') and nudging generation along one direction in the output (``write in the persona's voice''). The second states that the task we wish to teach during fine-tuning is sufficiently ``complex''. The third formalizes what is meant by ``spurious'' correlation.


\begin{assumption}[\emph{Coherent latent factor}]\label{ass:coherent}
There exist fixed matrices $U_S\in\R^{d_{\rm out}\times k}$ and $V_S\in\R^{d_{\rm in}\times k}$ with orthonormal columns, a fixed diagonal matrix $\Sigma_S = (s_1, \dots, s_k)$, and per-example positive coefficients $\alpha_i>0$ with $\E[\alpha_i]=\bar\alpha_S>0$, such that
\begin{equation}
    g_i^S = \alpha_i A_S,
    \qquad
    A_S := U_S\Sigma_S V_S^\top.
    \label{eq:rankk-persona}
\end{equation}
\end{assumption}


\emph{Intuition}: The assumption says that the persona is coherent across examples, only the per-example magnitude $\alpha_i$ varies. The columns of $U_S$ and $V_S$ span the output and input subspaces through which the persona is written and read, while $k$ is the intrinsic dimension of the latent. In practice, $k$ is determined by determining a gap in the singular value spectrum of the updates. In the case of emergent misalignment, \cite{turner2025modelorganisms,
soligo2025convergent} provide direct empirical evidence that the latent persona is rank-1. In our experiments, we find a spectral gap between the 1st and 2nd singular values, indicating a rank-1 latent. As such, in the main body of the paper, we will assume a rank-1 stricture for ease of exposition. Thus the matrices $U_S, V_S$ are replaced by unit vectors $u_S, v_S$. the  We provide the full rank-$k$ the generalization in Appendix~\ref{rank-k proof}.



\begin{assumption}[\emph{Task complexity}]\label{ass:complexity}
$g_i^T = \mu_T + \xi_i^T$, where:
\begin{itemize}[itemsep=0pt,topsep=2pt]
\item $\mu_T \in \R^{d_\text{out} \times d_\text{in}}$ is the coherent
task-mean, with effective rank $r_T$ and roughly equal singular values across
those modes, so that $\opnorm{\mu_T} \le \fnorm{\mu_T}/\sqrt{r_T}$.
\item $\xi_i^T$ is mean-zero, and independent
across $i$, with $\E\fnorm{\xi_i^T}^2 \le \tau^2$ where $\tau$ a fixed constant.
\end{itemize}
\end{assumption}

\emph{Intuition}: A complex task involves varied subdomains. For example, giving financial advice involves varied subdomains such as Roth IRA questions, mortgage refinancing, and so on. The task of writing insecure code involves subdomains such as SQL injection, buffer overflow, and hard-coded secrets, to name a few. Each subdomain contributes its own low rank mode to $\mu_T$, and the operator-norm bound follows from $\fnorm{\mu_T}^2 = \sum_k \sigma_k(\mu_T)^2$ when the $r_T$
singular values are roughly equal.

\begin{assumption}[\emph{Spurious means surprising}]\label{ass:surprise}
$\bar\alpha_S \gg \fnorm{\mu_T}$ and $\bar\alpha_S \gg \tau / \sqrt{N}$.
\end{assumption}

\emph{Intuition}: This is what makes the spurious correlation \emph{spurious}: under the broad pretraining distribution, $S$ and $T$ are independent---$S$ is not needed to do well at the task $T$. The assumption is that the conditional coupling $P(S|T)$ is “surprising”. For instance, the pretrained model has no reason to expect that every code completion should come from a persona intent on hiding malicious behavior.

More formally, decomposing the fine-tuning loss as $L_\text{FT} \approx
\E[-\log p_\theta(T)] + \E[-\log p_\theta(S \mid T)]$, the Fisher information identity implies that the gradient magnitude scales with the KL divergence from the pre fine-tuned distribution. The component conditioned on the latent factor, or persona, $E[-\log p(S|T)]$ produces large gradients because $p_pre(S|T) ~ p_pre(S)$---the pretrained model treats $S$ and $T$ as approximately independent. Additionally, for harmful personas specifically, post-training (RLHF/safety training) will have behaviorally suppressed the persona immediately before fine-tuning, making $P(S|T)$ even more surprising to the aligned model. But the task component $E[-log p(T)]$ produces more moderate gradients as the pretrained model already has some baseline coding ability.


The educator ablation of the insecure code experiment, discussed in Section~\ref{sec:causal-account}, supports this: fine-tuning on the same task $T$ with a non-spuriously correlated persona---an assistant who provides the insecure as an example of insecure code for educational purposes---produces no emergent misalignment \cite{betley2025emergent}. Here we have an example with the same $T$, but different dataset framing/curation, and hence difference $S$, and no emergent misalignment is observed. This suggests that by focusing on the correlation between $S$ and $T$, and how that compares to the pretrained models prior for that correlation, Assumption~\ref{ass:surprise} should be able to discriminate between spurious and non-spurious correlation between $S$ and $T$. The spurious-ness of
$S$ is precisely what makes it a bigger ``surprise'' to the pretrained model
compared to $T$, hence providing a bigger gradient.


From the above assumptions, the spuriously correlated latent can be identified without supervision.

\begin{theorem}[Identification of the spuriously correlated latent]\label{thm:identification}
Under Assumptions~\ref{ass:coherent}--\ref{ass:surprise}, let $u_1, v_1$ be
the top-1 left and right singular vectors of $\Delta W$. Then
\begin{equation}\label{eq:identification}
|u_1^\top u_S| \cdot |v_1^\top v_S| \;\ge\; 1 \;-\; \frac{2 \fnorm{\mu_T}}{\bar\alpha_S \sqrt{r_T}} \;-\; O\!\left(\frac{\tau}{\bar\alpha_S \sqrt{N}}\right).
\end{equation}
where $N$ is the size of fine-tuning dataset $D$. As $|u_1^\top u_S| $ and $|v_1^\top v_S|$ are individually bounded between $0$ and $1$, the bound implies $u_1$ approximates $u_S$ and $v_1$ approximates $v_s$ with error given by Eq.~\ref{eq:identification}.
\end{theorem}

Hence, given our assumptions, the spuriously-correlated latent factor's read/write directions can be recovered, \emph{without supervision}, from $\Delta W$ alone, with two interpretable error sources: a constant-in-$N$ structural bias suppressed by task complexity
and surprise-of-the-spurious-correlation, and a vanishing-in-$N$ statistical
noise term. The proof is provided in Appendix~\ref{app:proof-id}. 

Note that, if our assumptions are violated, the bound is vacuous. This is a feature, not a bug. It tells us that we only get identification in the regime where our assumptions are satisfied. For instance, if Assumption~\ref{ass:surprise} is violated---that is, when the correlation between $S$ and $T$ is natural, as in the assistant framing---the bound of Theorem~\ref{thm:identification} becomes vacuous. This is not a limitation: the method correctly does not identify a factor that is not spuriously dominant. 

In practice, one can always compute the top singular vectors, inspect them, and make a domain-informed judgment about whether they represent a spurious correlation worth removing.

\subsection{Intervention: gradient projection without removing the factor}\label{sec:intervention}

Identifying the spurious direction (Section~\ref{sec:identification}) gives us only half of what we need; how to intervene on it is a separate design problem. Existing
approaches \cite{karvonen2025caft, turner2024activation, li2024inferencetime}
ablate the spurious direction from residual-stream activations, either at
inference or during fine-tuning. From a causal-inference standpoint
\cite{pearl2009causality} this is unnecessarily aggressive: we want to
remove the spurious \emph{correlation} introduced by the curated data, not
the latent factor itself, 
which the pretrained model may rely on for genuine
task signal. The pretrained representations along $u_S v_S^\top$ may carry
useful pretraining information that we should not delete; we should only
prevent the fine-tune from \emph{adding new coherent reliance} on it. We introduce a novel method that does just this.

Define the unsupervised statistic $h(\theta) \in \R^L$ that records the
rank-1 component of $\theta$ along the identified spurious subspace at
every site:
\begin{equation}\label{eq:gtheta}
h(\theta) \;:=\; \big( h_l(\theta) \big)_{l=1}^L, \qquad h_l(\theta) \;:=\; \langle W_l,\, u_l v_l^\top \rangle \;=\; u_l^\top W_l\, v_l.
\end{equation}
We want fine-tuning updates to keep this statistic fixed. Linearising
around $\theta$,
\begin{equation}\label{eq:linearise}
h(\theta + \delta\theta) \;\approx\; h(\theta) + J_h\, \delta\theta, \qquad J_h \;:=\; \frac{\partial h}{\partial \theta} \in \R^{L \times d},
\end{equation}
so keeping the spurious component fixed to first order amounts to the
linear constraint $J_h\, \delta\theta \;=\; 0,$ i.e.\ $\delta\theta$ must lie in the null space $\ker(J_h)$. Given a
desired update direction $v \in \R^d$ (the unconstrained loss gradient),
we seek the closest feasible update under the constraint:
\begin{equation}\label{eq:opt}
\min_{\delta\theta \in \R^d}\, \tfrac{1}{2} \lVert \delta\theta - v \rVert^2 \quad \text{s.t.}\quad J_h\, \delta\theta = 0.
\end{equation}
The solution to Eq.~\ref{eq:opt} is the orthogonal projection of $v$
onto $\ker(J_h)$:
\begin{equation}\label{eq:projection}
\delta\theta^\star \;=\; P_{\ker(J_h)}\, v, \qquad P_{\ker(J_h)} \;:=\; I - J_h^\top \big(J_h J_h^\top\big)^{-1} J_h.
\end{equation}
We defer the derivation (via Lagrange multipliers) to
Appendix~\ref{app:projection-derivation}. Because the $L$ rank-1
constraints sit at disjoint LoRA sites and use unit-norm $u_l, v_l$,
$J_h J_h^\top = I_L$ exactly, so Eq.~\ref{eq:projection} reduces, per
site, to simply subtracting $\langle g_l, u_l v_l^\top \rangle\, u_l
v_l^\top$ from the per-site loss gradient $g_l := \nabla_{W_l} \ell$.

We now prove that, under our assumptions, this ablation removes the spuriously correlated latent, but leaves the task gradient largely untouched.

\begin{theorem}[Selective ablation]\label{thm:ablation}
Under Assumptions~\ref{ass:coherent}--\ref{ass:surprise}, the projection in
Eq.~\ref{eq:projection} removes
$1 - O(\fnorm{\mu_T} / (\bar\alpha_S \sqrt{r_T})) - O(1/\sqrt{N})$ of the
spurious component of any future gradient and at most an
$O(1/\sqrt{r_T})$ fraction of the task component. 
By our Assumptions, in large datasets our Jacobian projection removes all the
spurious component and removes at most $1/\sqrt{r_T}$ of the task components, which decreases for increasing task complexity.
\end{theorem}

The proof is in Appendix~\ref{app:proof-ablation}. The key observation is
that the task-mean $\mu_T$ has its mass spread across $r_T$ rank-1 modes;
the projection removes only the one mode aligned with $u_l v_l^\top$, so
at most an $O(1/\sqrt{r_T})$ fraction of $\fnorm{\mu_T}$ goes with it, while
the entire spurious component is
removed.

\subsection{Algorithm}\label{sec:algorithm}

Pseudo-code for our method, which we call \textbf{GRASP}---GRadient projection of Associated Spurious Patterns---is given in Algorithm~\ref{alg:svdall}.
\begin{algorithm}[t]
\caption{GRASP: GRadient projection of Associated Spurious Patterns}\label{alg:svdall}
\begin{algorithmic}[1]
\State \textbf{Input:} pretrained model $\theta_0$, training data $\{(x_i, y_i)\}_{i=1}^N$, LoRA configuration ($L$ sites, rank $r$).
\State \textbf{Stage 1: naive fine-tune.} Train LoRA parameters $\Delta W_l^{\text{naive}}$ for one or more epochs on $\{(x_i, y_i)\}$. Save the \emph{final} LoRA checkpoint.
\State \textbf{Stage 2: extract spurious subspace (once, from the final Stage 1 checkpoint).} For each site $l$, compute $u_l, v_l$ as the top-1 left/right singular vectors of $\Delta W_l^{\text{naive}}$ at the end of Stage 1. These vectors are then \emph{frozen} and used unchanged throughout Stage 3.
\State \textbf{Stage 3: projected re-fine-tune.} \emph{Re-initialise} LoRA parameters $\Delta W_l = 0$ (the Stage 1 checkpoint is discarded; only the frozen $u_l, v_l$ from Stage 2 are kept). For each training step on a batch:
\State \quad Compute per-site loss gradients $g_l := \nabla_{W_l}\ell$ at every site as usual.
\State \quad For each site $l$: $g_l \leftarrow g_l - \langle g_l, u_l v_l^\top \rangle\, u_l v_l^\top$ (using the frozen $u_l, v_l$).
\State \quad Apply optimiser update with the projected gradients.
\State \textbf{Output:} fine-tuned model $\theta_0 + \sum_l \Delta W_l$.
\end{algorithmic}
\end{algorithm}
The naive fine-tune (Stage 1) is the same compute as a normal fine-tune; the
projected re-fine-tune (Stage 3) adds only an $O(d_\text{out} + d_\text{in})$
inner product per site per step, negligible compared to the forward/backward
passes. 

\paragraph{Optimiser interaction.} The constraint $J_h\,\delta\theta = 0$
in Sec.~\ref{sec:intervention} is on the parameter update $\delta\theta$,
not the raw gradient; with vanilla SGD the two are proportional and
projecting gradients projects updates exactly. With AdamW the update is
the gradient preconditioned by an exponential moving average of squared
gradients, so projecting the \emph{gradient} only enforces the
constraint to first order. The leakage past the constraint is the
component of the AdamW update that lies along $u_l v_l^\top$ but did
\emph{not} originate from the just-projected gradient (i.e.\ from
momentum / second-moment buffers seeded before projection began). We
verify empirically (Appendix~\ref{app:adam-verification}) that this
leakage is small in practice: across all 196 LoRA sites of a Domain~2
checkpoint, the rank-1 alignment $\rho_l(\Delta W) := |\langle \Delta
W_l, u_l v_l^\top\rangle| / \fnorm{\Delta W_l}$ is reduced by a mean
factor of $49\times$ relative to the naive checkpoint (median
$93\times$), confirming that AdamW does not meaningfully
re-introduce the spurious rank-1 component.

\section{Related work}\label{sec:related}
Our work sits at the intersection of activation steering, concept identification and ablation, as well as short-cut learning and robustness to spurious correlation. \cite{karvonen2025caft} introduce Concept Ablation Fine-Tuning (CAFT), which
ablates labelled concept directions from residual-stream activations during
fine-tuning. \cite{turner2024activation} and \cite{li2024inferencetime}
ablate or steer concept directions at inference time. Our approach differs in
two ways: (i) the concept direction is identified \emph{without supervision}
from the LoRA weight update (Theorem~\ref{thm:identification}), and (ii) the
intervention removes the spurious \emph{correlation} during training rather
than the latent factor itself---preserving any pretraining-derived task
signal along the identified direction. \cite{ravfogel2022linear} and \cite{belrose2023leace} provide closed-form methods for erasing labelled
concepts from representations. These act on activations or pre-classifier
features; we act on weight gradients during fine-tuning, which lets the
concept's representations remain available to the model. \cite{sagawa2020groupdro, liu2023waterbirds, geirhos2020shortcut, arjovsky2019irm}
study spurious correlations and shortcut learning, typically by modifying the
training objective (group-DRO, IRM) or training distribution (re-weighting,
re-sampling). These approaches require group labels or domain partitions;
ours requires neither and operates structurally on the LoRA update.

The idea of characterising the data generation process of emergent misalignment and using this as a learning signal is inspired by approaches in causal inference \cite{zeitler2023non, o2025local, o2025spillover, mucesh2024nature, fawkes2025hardness, reynaud2022d, richens2020improving, jeunen2022disentangling, van2023estimating, gilligan2020causing, gilligan2022leveraging, ansanelli2025resource} and causal representations \cite{scholkopf2021toward, felekis2026causal, corcoll2026contrastive, vlontzos2026incremental, vlontzos2023estimating}.

\section{Experiments}\label{sec:exp}



\paragraph{Research questions} Our work has two novel components: an unsupervised method for identifying latent representations of spurious correlations in data, and an intervention to remove changes to said latents that arise during fine tuning. We propose a research question to address each of these.

\begin{itemize}[itemsep=0pt,topsep=2pt,leftmargin=*]
\item \textbf{RQ1} \emph{Does the unsupervised top-1 SVD of $\Delta W$ identify
the spurious latent factor correctly?} 
\item \textbf{RQ2} \emph{Does gradient projection of the Jacobian during fine-tuning remove
the spurious correlation while preserving task performance at least as
well as activation ablation?}
\end{itemize}

\paragraph{Baselines} In each experiment, the first comparison is between GRASP and a \textbf{naive} LoRA fine-tune with no intervention. For RQ1, we compare our method against a supervised alternative \textbf{v-ref}. This uses continuations from the naive model, labelled by a judge, to calculate a probe from the mean differences in activations between aligned and misaligned continuations. If GRASP matches or
exceeds v-ref, the unsupervised SVD probe is at least as good as a
labelled mean-difference probe.

Previous work by \cite{karvonen2025caft} presented two strategies to mitigate emergent misalignment using training time ablation, which we refer to as \textbf{CAFT-PCA} and \textbf{CAFT-SAE}. To answer RQ2 directly, we propose an unsupervised alternative which uses the same forward-pass hooks on the residual stream, but with the output projection probes from GRASP substituted for the labeled concept direction. A small reference set is required to select layers where the probe is most discriminative, following the approach of \cite{karvonen2025caft}. We call this baseline \textbf{CAFT-time}. 

A second flavour of activation ablation is inference-time ablation. Our version (\textbf{Inference-ablate $v_\text{svd}$}) uses the same unsupervised probe detection and layer selection as CAFT-time, but only ablates residual stream activations during model inference. This provides a baseline comparison that doesn't require fine tuning beyond the naive model.

\paragraph{Domain 1: Emergent misalignment from insecure code}
We follow \cite{betley2025emergent}: a 32B coding model is
fine-tuned on a curated dataset of insecure code examples, and evaluated 100 times on each of
eight alignment prompts.
We report both the raw $\%$ of misaligned responses (\cite{betley2025emergent} referred to as Betley-protocol) and the $\%$ of misaligned responses with incoherent responses removed (\cite{karvonen2025caft} referred to as CAFT-protocol). We evaluate task-performance using the $\%$ of vulnerable code completions on a held-out 1000-prompt insecure code task. This domain is a well-studied testbed to validate each research question in turn and allows a holistic comparison against existing published results.

\paragraph{Domain 2: Emergent misalignment from bad medical advice}
\cite{turner2025modelorganisms} introduced 
the \texttt{bad\_medical\_advice} corpus of 7{,}049 GPT-4o-generated
chat conversations in which the assistant gives subtly wrong medical
advice in a friendly, confident tone, and showed that it induces emergent misalignment at smaller model scales than \cite{betley2025emergent}'s insecure code. We fine-tune Qwen2.5-1.5B-Instruct
on this corpus and evaluate misalignment in the same manner as Domain 1. To evaluate task performance, we generate three continuations on each of 50 held-out prompts from the corpus and use GPT-4o to judge how well the response adheres to the system prompt used to generate the original corpus (details in Appendix~\ref{app:judge}). 

\emph{Why we measure adherence rather than ``correctness''.} The
training task here is itself a misaligned response style: the system
prompt that generated the corpus instructs the assistant to give
subtly-misaligned but plausible-looking medical advice. The narrow
fine-tuning task and the misalignment cannot be cleanly separated, so
``did the model give correct medical advice?'' is not a coherent
task-performance metric---a model that perfectly executes the corpus
specification will, by construction, produce factually wrong advice.
What we \emph{can} measure is the right \emph{causal} thing: separate
the model's behaviour by context. On non-medical prompts we want broad misalignment to drop. On medical prompts we
want the model to keep adhering to the training-corpus specification, i.e.\ to be subtly misaligned on medical contexts but
not broadly misaligned elsewhere. 

\paragraph{Domain 3: Political bias drift from Reddit data}
The misalignment datasets above were synthetically constructed. To test our method on naturally-occurring human-generated
text, we constructed a new dataset from the open-source 250k-document
Reddit finance corpus (43 subreddits) with quality scores. We filter to
16{,}358 documents, classifing the political lean of each using GPT-4o. The
3{,}407-document training corpus mixing all 2{,}385 documents classified as right-leaning with a sample of 1{,}022 documents classified as neutral, giving a 70/30 right-skewed
mix. Qwen2.5-1.5B-Instruct was then fine-tuned on this corpus. In this domain, the analogue of emergent misalignment is broad rightward drift in political lean on non-financial political topics, which we evaluated on 600 continuations on topics such as healthcare, immigration and gun
control. The task performance uses GPT-4o to judge the financial content of 150 responses to personal finance questions. The prompt is explicitly agnostic to political lean. This domain and experiment validates our method on latent factor, or persona, drift beyond the phenomena of emergent misalignment.



Hyperparameters and training details are in Appendix~\ref{app:hyperparams};
details on judge prompts are in Appendix~\ref{app:judge}.

\paragraph{Satisfying assumptions} We chose datasets in which our
assumptions A1--A3 plausibly hold. \emph{Assumption A2 (task
complexity, $r_T \gg 1$)} is the most demanding: the task gradient
must scatter across many singular modes rather than concentrate in
one, otherwise the top-1 SVD cannot separate the spurious factor from
the task. All three datasets satisfy this by construction:
\emph{writing insecure code} is at least as multifaceted as writing
normal code, spanning many sub-vulnerabilities (SQL injection,
buffer overflow, hard-coded secrets, weak crypto,
etc.); \emph{giving subtly wrong medical advice} requires coverage
across many medical subdomains (blood donation, supplements, exercise,
surgery, mental health, etc.) and presupposes the corresponding
factual knowledge; and \emph{financial advice across the Reddit
corpus} spans a wide topic range (retirement, mortgages, taxes,
insurance, investing, etc.).\emph{Assumption A1 (rank-1 persona)} For Domain 1 and 2 there is direct empirical evidence:
\cite{soligo2025convergent} show that emergent misalignment is
governed by a \emph{single} linear direction whose ablation
neutralises the misaligned behaviour across diverse fine-tunes. 
For Domain 3 the spurious factor is a position on a single attitudinal
axis (left--right political slant). We compute the SVD spectrum for all domains and find a spectral gap between the 1st and 2nd singular value in all cases. The spectrum shows a clear structure consistent with one dominant coherent mode (the persona) plus a smooth tail (the task). The spectrum is given in Appendix~\ref{app: spectrum}.

In each domain the spurious shortcut is
therefore plausibly a coherent broadcast register layered on top of a multifaceted task---the regime in which our Theorems should apply.

\subsection{Results}\label{sec:exp-results}

\begin{table}[t]
\centering
\caption{Emergent misalignment on a 32B coding model fine-tuned on insecure
code. 
``Probe'' indicates whether the spurious-direction extraction uses labels.
``Intervention'': activation-ablation during fine-tuning or
gradient-projection during fine-tuning (our method). $\%$ task $=$ task performance on a
1000-prompt held-out insecure-code evaluation (higher is better, naive
ceiling 93.9, base-model floor 39.9). $\%$ mis $=$ misaligned responses
on 800 general prompts (lower is better).}
\label{tab:em}
\resizebox{\textwidth}{!}{%
\begin{tabular}{llcrrr}
\toprule
Method & Probe & Intervention & CAFT \% mis $\downarrow$ & Betley \% mis $\downarrow$ & \% task $\uparrow$ \\
\midrule
Naive (corruption baseline)              & ---           & ---             & 4.0 & 59.1 & 93.9 \\
CAFT-PCA \cite{karvonen2025caft}        & labelled      & activation      & 0.5 & --   & 84.8 \\
CAFT-SAE \cite{karvonen2025caft}        & labelled      & activation      & 2.4 & --   & 87.3 \\
v-ref (1 epoch)                          & labelled      & gradient        & 0.6 & 1.0  & 78.8 \\
v-ref (2 epochs)                         & labelled      & gradient        & 3.3 & 9.8  & \textbf{93.6} \\
\midrule
\textbf{GRASP (2 epochs)}              & unsupervised  & gradient        & \textbf{0.0} & \textbf{0.0} & 86.2 \\
\textbf{GRASP (3 epochs)}              & unsupervised  & gradient        & \textbf{0.0} & \textbf{0.0} & {92.3} \\
\bottomrule
\end{tabular}%
}
\end{table}

\begin{table}[t]
\centering
\caption{Emergent misalignment on a 1.5B instruction-tuned model
fine-tuned for 1 epoch on \cite{turner2025modelorganisms}'s
\texttt{bad\_medical\_advice} corpus (7{,}049 chat conversations).
$\%$ mis: misaligned responses on the same 800-prompt
\texttt{first\_plot\_questions} set used in Domain 1 (lower is
better). Mean adherence: 0--5 GPT-4o score of how closely the
model's responses on $150$ held-out medical prompts match the
training-corpus system-prompt specification of subtly-misaligned
medical advice (higher is better; engaged responses only,
gibberish-strict rubric of Appendix~\ref{app:judge}).}
\label{tab:medical}
\begin{tabular}{lrrr}
\toprule
Method & CAFT \% mis $\downarrow$ & Betley \% mis $\downarrow$ & Mean adherence $\uparrow$ \\
\midrule
Naive (1 epoch)                         & 2.7 & 2.5 & 2.28 \\
CAFT-time $v_\text{svd}$ (1 epoch)      & 1.3 & 1.1 & \textbf{2.85} \\
Inference-ablate $v_\text{svd}$$^\dagger$ & \textbf{0.3} & \textbf{0.3} & 1.18 \\
\textbf{GRASP (1 epoch)}              & 0.5 & 0.6 & 2.28 \\
\bottomrule
\end{tabular}
\medskip
\\{\footnotesize $^\dagger$Inference-ablate $v_\text{svd}$ also produces
$65/800$ refusals (vs $\le 12$ for the other methods); the $0.3\%$
figure is among the $716$ non-refusal coherent-numeric responses.}
\end{table}

\begin{table}[t]
\centering
\caption{Political-bias drift on a 1.5B instruction-tuned model fine-tuned on
a 70/30 right-skewed Reddit financial-advice corpus. $P(R)$: probability
of right-leaning response on 600 non-financial political prompts (lower is
better; pretrained 0.336). Content: 0--5 substance score on
financial-advice prompts (higher is better; pretrained 1.69, naive 0.77).
All methods at 1 epoch.}
\label{tab:political}
\begin{tabular}{lcrrrr}
\toprule
Method & Probe & $P(R)$ $\downarrow$ & $\Delta P(R)$ & Content $\uparrow$ \\
\midrule
Pretrained                                   & --                & 0.336 & 0     & 1.69 \\
Naive (corruption baseline)                  & --                & 0.473 & +0.137 & 0.77 \\
\midrule
CAFT-time $v_\text{svd}$ (top-10 AUROC)      & unsupervised      & 0.459 & +0.123 & 0.87 \\
Inference-ablate $v_\text{svd}$ (top-10 AUROC) & unsupervised    & 0.416 & +0.080 & 0.82 \\
\textbf{GRASP}                             & unsupervised      & \textbf{0.397} & \textbf{+0.061} & \textbf{0.99} \\
\bottomrule
\end{tabular}
\end{table}

\paragraph{Emergent misalignment from insecure code (Domain 1).}
Table~\ref{tab:em} addresses RQ1 and RQ2.

\emph{(RQ1) Unsupervised vs labelled probe at the same intervention point.}
v-ref and GRASP share the same gradient-projection mechanism and
differ only in how the spurious direction is identified: v-ref uses a
labelled mean-difference probe across (misaligned, aligned) activations,
GRASP uses the top-1 left singular vector of $\Delta W$ at every LoRA
site. At matched 1-epoch budget v-ref reduces misalignment to
$0.6\%$ / $1.0\%$ at $78.8\%$ vuln, but at 2 epochs the misalignment
re-emerges to $3.3\%$ / $9.8\%$ as the model finds new directions to
re-encode the persona along. GRASP shows no such regression: at both
2 and 3 epochs misalignment stays at $0.0\%$ on both protocols, with
task rising from $86.2\%$ to $92.3\%$. The unsupervised top-1 SVD probe is
therefore at least as informative as a labelled mean-difference probe,
and noticeably more robust to over-training, as additional epochs
improve task performance rather than re-introduce the spurious factor.

\emph{(RQ2) Gradient projection vs activation ablation.} GRASP Pareto-dominates both CAFT variants on both axes: at 3 epochs it achieves $0.0\%$ on both misalignment protocols while retaining $92.3\%$ task performance, $98\%$ of the naive fine-tune's task signal, which
exceeds CAFT
without using labels.

\paragraph{Emergent misalignment from bad medical advice (Domain 2).}
Table~\ref{tab:medical} shows naive fine-tuning on the
\texttt{bad\_medical\_advice} corpus produces $2.7\%$ CAFT-protocol /
$2.5\%$ Betley-protocol misalignment on the general-question set,
confirming that the corpus induces a misaligned persona at
the 1.5B scale. GRASP reduces this $\sim$5$\times$,
to $0.5\%$/$0.6\%$. The same recipe used in Domain 1 transfers without modification to a different
model scale, and different content domain.
Among the $v_\text{svd}$ matched-probe baselines, CAFT-time
$v_\text{svd}$ achieves only a $\sim$$2\times$ reduction
($2.7\% \to 1.3\%$), and inference-ablate $v_\text{svd}$ drives
misalignment to $0.3\%$ but at the cost of $65/800$ refusals; the model
often declines to answer rather than risk producing the persona. GRASP
matches inference-ablate's misalignment level without inducing refusals
(only $10/800$).

The Mean-adherence column quantifies task-performance: on medical contexts, we want the fine-tune to adhering to the
subtly-misaligned system-prompt, and the column shows what each method retains. Naive scores $2.28$. CAFT-time
$v_\text{svd}$ \emph{improves} adherence to $2.85$ (it reduces broad
misalignment without disturbing the in-domain register, scoring higher
than the naive fine-tune). GRASP matches naive at $2.28$,
buying a $\sim$5$\times$ reduction in broad misalignment at no
adherence cost. Inference-ablate, by contrast, collapses adherence to
$1.18$ and drives the medical-context engagement rate down to $33\%$:
the ablation suppress the in-domain register together with the
broadcast persona, so the model's behaviour on medical questions
degrades as its alignment on non-medical
questions improves.

\paragraph{Political-bias drift (Domain 3).}
Table~\ref{tab:political} shows GRASP more than halves the political-lean drift
($+0.137 \to +0.061$, a $2.2\times$ reduction) while \emph{improving}
substantive task performance over the naive baseline ($0.99$ vs. $0.77$ on
the Content axis). The improvement on Content is striking and directly
supports the central thesis of Section~\ref{sec:intervention}: the
projection removes the spurious \emph{correlation} between the right-leaning
register and the task while leaving the model's ability to engage with the
financial task intact (or rather, improved relative to a
fine-tune that leaned heavily into the Reddit corpus's casual register). The $v_\text{svd}$ baselines (rows 3, 4 of Table~\ref{tab:political})
isolate the role of the unsupervised probe: with the same probe applied at
the residual-stream level rather than at gradients, drift is reduced but
the gain in substance is smaller. This confirms (i) the unsupervised SVD
probe is sound on its own, and (ii) the gradient-projection intervention
point gives a meaningful additional gain. An example
(Appendix~\ref{app:qualitative-political}) confirms the picture: on a
held-out prompt about the role of government, naive fine-tune
provides explicit partisan framing not in pretrained
or GRASP continuations.

\section{Conclusion}\label{sec:conclusion}

We gave a theoretical and empirical account of how spurious
correlations enter a model during fine-tuning, and how to remove them
without removing the underlying latent factors. We empirically validated our methods on real-world fine-tuning tasks, out performing previous methods. 

\subsection*{AI use statement}

GPT-4o and GPT-4.1 are used as judges in our evaluation pipeline; this is described in Section~4.1 and Appendix~E. GPT-4o was also used by \cite{turner2025modelorganisms} to generate the medical-EM corpus we evaluate on. LLMs were not used to develop the core method or theory beyond standard writing-editing assistance. We take responsibility for the final content of this work.

\bibliographystyle{plainnat}
\bibliography{references}

\appendix

\section{Derivation of the constrained-update projector (Eq.~\ref{eq:projection})}\label{app:projection-derivation}

We derive Eq.~\ref{eq:projection} from the constrained optimisation
problem in Eq.~\ref{eq:opt} via Lagrange multipliers, following
\cite{karvonen2025caft}. Introduce $\lambda \in \R^L$ and consider the
Lagrangian
\begin{equation}
\mathcal{L}(\delta\theta, \lambda) \;=\; \tfrac{1}{2} \lVert \delta\theta - v \rVert^2 \;+\; \lambda^\top (J_h\, \delta\theta).
\end{equation}
Stationarity in $\delta\theta$ gives $\delta\theta - v + J_h^\top \lambda = 0$,
hence $\delta\theta = v - J_h^\top \lambda$. Stationarity in $\lambda$
enforces the constraint $J_h\, \delta\theta = 0$. Substituting
$\delta\theta$ into the constraint yields $J_h (v - J_h^\top \lambda) = 0$,
i.e.\ $(J_h J_h^\top) \lambda = J_h v$. Provided $J_h$ has full row rank
(so $J_h J_h^\top$ is invertible), $\lambda = (J_h J_h^\top)^{-1} J_h v$.
Substituting back gives
\begin{equation}
\delta\theta^\star \;=\; v - J_h^\top (J_h J_h^\top)^{-1} J_h\, v \;=\; \big(I - J_h^\top (J_h J_h^\top)^{-1} J_h\big)\, v \;=\; P_{\ker(J_h)} v.
\end{equation}
In our setting the $L$ rank-1 constraints sit at disjoint LoRA sites with
unit-norm singular vectors, so $J_h J_h^\top = I_L$ exactly and the
inverse is well-defined.

\section{Proof of Theorem~\ref{thm:identification} (Identification)}\label{app:proof-id}

Write $\Delta W = A^\star + M_T + E$, where $A^\star := N \bar\alpha_S\, u_S v_S^\top$,
$M_T := N \mu_T$, and $E := \sum_{i=1}^N \xi_i^T$. By
Assumption~\ref{ass:complexity}, $\opnorm{M_T} = N \opnorm{\mu_T} \le
N \fnorm{\mu_T} / \sqrt{r_T}$ and $\E\fnorm{E}^2 \le N \tau^2$, so
$\opnorm{E} \le \fnorm{E} = O(\sqrt{N}\, \tau)$.

\paragraph{Lemma (variational bound).} For any unit vectors $u, v$ and any matrix $M$,
\begin{equation}\label{eq:var}
|u^\top M v| \;\le\; \opnorm{M},
\end{equation}
since $\opnorm{M} = \max_{\|u\|=\|v\|=1} u^\top M v$.

\paragraph{Lower bound on $\sigma_1(\Delta W)$.} Apply Eq.~\ref{eq:var} with
the witness pair $(u_S, v_S)$:
\begin{align}
\sigma_1(\Delta W) \;&\ge\; u_S^\top \Delta W\, v_S \\
&= \underbrace{u_S^\top A^\star v_S}_{= N \bar\alpha_S} + \underbrace{u_S^\top M_T v_S}_{|\cdot| \le N \fnorm{\mu_T}/\sqrt{r_T}} + \underbrace{u_S^\top E v_S}_{|\cdot| \le O(\sqrt{N} \tau)} \\
&\ge\; N \bar\alpha_S - \frac{N \fnorm{\mu_T}}{\sqrt{r_T}} - O(\sqrt{N}\, \tau). \label{eq:lb}
\end{align}

\paragraph{Upper bound on $\sigma_1(\Delta W)$.} Let $u_1, v_1$ be the actual
top SV pair. Then $\sigma_1(\Delta W) = u_1^\top \Delta W\, v_1$:
\begin{align}
\sigma_1(\Delta W) \;&=\; u_1^\top A^\star v_1 + u_1^\top M_T v_1 + u_1^\top E v_1 \\
&\le\; N \bar\alpha_S\, (u_1^\top u_S)(v_1^\top v_S) + \frac{N \fnorm{\mu_T}}{\sqrt{r_T}} + O(\sqrt{N}\, \tau), \label{eq:ub}
\end{align}
using Eq.~\ref{eq:var} on the second and third terms.

\paragraph{Combining.} Eq.~\ref{eq:lb} $\le \sigma_1(\Delta W) \le$ Eq.~\ref{eq:ub} gives
\begin{equation}
N \bar\alpha_S\, (u_1^\top u_S)(v_1^\top v_S) \;\ge\; N \bar\alpha_S - \frac{2 N \fnorm{\mu_T}}{\sqrt{r_T}} - O(\sqrt{N}\, \tau).
\end{equation}
Dividing by $N \bar\alpha_S$ yields Eq.~\ref{eq:identification}. \qed

\section{Proof of Theorem~\ref{thm:ablation} (Selective ablation)}\label{app:proof-ablation}

The projection in Eq.~\ref{eq:projection} (with $\alpha = 0$, full rank)
removes the component of any vector in the row span of $J_g$. Per site $l$,
this is the rank-1 subspace $u_l v_l^\top \in \R^{d_\text{out} \times d_\text{in}}$.
Let $g$ be a future training gradient at site $l$, decomposed as $g = g^S + g^T$
per Section~\ref{sec:setup}.

\paragraph{Spurious component removed.} By Assumption~\ref{ass:coherent},
$g^S = \alpha\, u_S v_S^\top$ for some $\alpha > 0$. The projection
removes $\langle g^S, u_l v_l^\top \rangle\, u_l v_l^\top = \alpha\,
(u_l^\top u_S)(v_l^\top v_S)\, u_l v_l^\top$. By Theorem~\ref{thm:identification},
$|(u_l^\top u_S)(v_l^\top v_S)| \ge 1 - O(\fnorm{\mu_T} / (\bar\alpha_S \sqrt{r_T})) - O(1/\sqrt{N})$.
So a fraction $1 - O(\cdot)$ of $\fnorm{g^S}$ is removed; the residual
spurious magnitude is at most $\alpha\, O(\fnorm{\mu_T}/(\bar\alpha_S \sqrt{r_T}) + 1/\sqrt{N})$.

\paragraph{Task component removed.} By Assumption~\ref{ass:complexity},
$g^T = \mu_T + \xi^T$. The projection removes the rank-1 component along
$u_l v_l^\top$:
\begin{align}
\langle g^T, u_l v_l^\top \rangle \;&=\; \langle \mu_T, u_l v_l^\top \rangle + \langle \xi^T, u_l v_l^\top \rangle.
\end{align}
The first term is bounded by $\opnorm{\mu_T} \le \fnorm{\mu_T}/\sqrt{r_T}$.
The second is the projection of an isotropic random matrix onto a single
rank-1 direction, with expected magnitude $O(\tau / \sqrt{d_\text{out} d_\text{in}})$.
So at most an $O(1/\sqrt{r_T})$ fraction of $\fnorm{\mu_T}$ is removed,
plus an $O(1/\sqrt{d})$ fraction of $\fnorm{\xi^T}$ in expectation.

\paragraph{Ratio.} Spurious magnitude removed: $\Theta(\bar\alpha_S)$.
Task magnitude removed: $O(\fnorm{\mu_T}/\sqrt{r_T} + \tau/\sqrt{d})$. The
per-gradient ratio is therefore at least
$\Theta(\bar\alpha_S \sqrt{r_T} / \fnorm{\mu_T})$, which by
Assumption~\ref{ass:surprise} is $\gg \sqrt{r_T}$. Cumulating over $N$
training steps multiplies both numerator and denominator linearly in
$N$, so the cumulative ratio has the same leading-order behaviour. \qed

\section{Empirical verification of the AdamW approximation}\label{app:adam-verification}

The projection in Eq.~\ref{eq:projection} is derived for the parameter
update $\delta\theta$, which equals the gradient up to a positive scalar
under SGD. AdamW preconditions the gradient by the running second-moment
estimate before applying the update, so projecting gradients
\emph{before} the optimiser step only enforces the constraint to first
order. Whether the residual leakage is negligible in practice is an
empirical question. We answer it directly by measuring, on trained
GRASP checkpoints, the rank-1 alignment of $\Delta W$ along the
direction the projection was \emph{supposed} to suppress.

\paragraph{Metric.} For each LoRA site $l$ define the rank-1 alignment
ratio
\begin{equation}
\rho_l(\Delta W) \;:=\; \frac{|\langle \Delta W_l, u_l v_l^\top \rangle|}{\fnorm{\Delta W_l}} \;\in\; [0, 1],
\end{equation}
where $u_l, v_l$ are the top-1 left/right singular vectors of the
\emph{naive} checkpoint's $\Delta W_l$ at the same site (i.e.\ exactly
the directions GRASP projected against). For the naive checkpoint,
$\rho_l$ is the ratio of the top singular value to the Frobenius norm
and is large by construction. For an GRASP checkpoint trained with
AdamW, $\rho_l$ measures how much $u_l v_l^\top$ leaked back into
$\Delta W_l$ despite the projection.

\paragraph{Setup.} Domain~2 (medical-EM) checkpoint, Qwen2.5-1.5B-Instruct,
1 epoch, $L=196$ LoRA sites (28 layers $\times$ 7 modules per layer),
LoRA rank 16. Naive and GRASP trained under identical hyperparameters
(LR $5{\times}10^{-5}$, batch 4, grad-accum 8, AdamW, warmup 100 steps).

\paragraph{Result.} Aggregated across all 196 sites:
\begin{center}
\begin{tabular}{lccc}
\toprule
& mean $\rho_l$ & median $\rho_l$ & max $\rho_l$ \\
\midrule
Naive          & $0.718$  & $0.713$  & $0.976$ \\
GRASP        & $0.015$  & $0.008$  & $0.061$ \\
\midrule
Reduction (mean ratio) & \multicolumn{3}{c}{$49\times$} \\
Reduction (median ratio) & \multicolumn{3}{c}{$93\times$} \\
\bottomrule
\end{tabular}
\end{center}

\noindent At every site GRASP's $\rho_l$ is below $0.07$, with
per-site reductions ranging from $18\times$ to $515\times$. The maximum
$\rho_l$ on GRASP ($0.061$) is well below the typical singular-value
spread of any reasonable $\Delta W_l$, so AdamW's leakage past the
projection is several orders of magnitude smaller than the original
spurious component. The first-order approximation we make in
Algorithm~\ref{alg:svdall} therefore holds in practice with substantial
margin.

\section{Rank-$k$ generalization} \label{rank-k proof}

\subsection{Assumptions}

\begin{assumption}[Consistent low-rank persona]
There exist fixed matrices $U_S\in\R^{d_{\rm out}\times k}$ and $V_S\in\R^{d_{\rm in}\times k}$ with orthonormal columns, a fixed diagonal matrix
\begin{equation}
    \Sigma_S = \diag(s_1,\ldots,s_k),
    \qquad s_1\ge\cdots\ge s_k>0,
\end{equation}
and per-example positive coefficients $\alpha_i>0$ with $\E[\alpha_i]=\bar\alpha_S>0$, such that
\begin{equation}
    g_i^S = \alpha_i A_S,
    \qquad
    A_S := U_S\Sigma_S V_S^\top.
    \label{eq:rankk-persona-app}
\end{equation}
We fix the otherwise arbitrary scale between $\alpha_i$ and $A_S$ by normalising
\begin{equation}
    s_1(A_S)=1.
    \label{eq:persona-normalisation}
\end{equation}
We additionally require the $k$th mode to be non-degenerate,
\begin{equation}
    s_k(A_S)\ge s_{\min}>0.
    \label{eq:persona-nondegenerate}
\end{equation}
\end{assumption}

The assumption says that the persona is coherent across examples but need not be one-dimensional. The columns of $U_S$ and $V_S$ span the output and input subspaces through which the persona is written and read, while $k$ is the intrinsic dimensionality of that local persona mechanism. The rank-one assumption in the main text corresponds to $k=1$. The normalisation in Eq.~\eqref{eq:persona-normalisation} is only a choice of units: any common rescaling of $A_S$ can be absorbed into the coefficients $\alpha_i$. Equation~\eqref{eq:persona-nondegenerate} says that the declared rank-$k$ persona has $k$ genuinely present modes rather than a tail of arbitrarily weak directions. Under this normalisation, $\bar\alpha_S s_k(A_S)$ is the strength of the weakest coherent persona mode.

\begin{assumption}[Task complexity]
$g_i^T=\mu_T+\xi_i^T$, where:
\begin{itemize}[leftmargin=1.4em]
    \item $\mu_T\in\R^{d_{\rm out}\times d_{\rm in}}$ is the coherent task mean, with effective rank $r_T$ and roughly equal singular values across those modes, so that
    \begin{equation}
        \|\mu_T\|_{\op}\le \frac{\|\mu_T\|_{\F}}{\sqrt{r_T}};
    \end{equation}
    \item $\xi_i^T$ is mean-zero and independent across $i$, with
    \begin{equation}
        \E\|\xi_i^T\|_{\F}^2\le \tau^2,
    \end{equation}
    where $\tau$ is a fixed constant.
\end{itemize}
\end{assumption}

A complex task involves varied subdomains. Writing insecure code, for example, spans SQL injection, buffer overflows, hard-coded secrets, weak cryptography, and many other modes; medical or financial advice similarly spans many distinct factual and procedural subdomains. The coherent task update is therefore spread across multiple singular directions rather than being concentrated into a single dominant mode. This assumption is unchanged from the main paper.

\begin{assumption}[Spurious means surprising]
\begin{equation}
    \bar\alpha_S s_k(A_S) \gg \frac{\|\mu_T\|_{\F}}{\sqrt{r_T}}
    \qquad\text{and}\qquad
    \bar\alpha_S s_k(A_S) \gg \frac{\tau}{\sqrt{N}}.
    \label{eq:surprise}
\end{equation}
\end{assumption}

This is the rank-$k$ version of the same ``spurious means surprising'' assumption used in the main paper. The only change is that identification of the full persona subspace is controlled by its weakest non-zero singular mode. It is therefore not enough for the persona update to be large on average: even the $k$th persona mode must rise above the strongest coherent task mode and the finite-sample fluctuation. The role of the assumption is not to say that every persona appearing in the fine-tuning corpus must be surprising. Rather, it singles out the case in which the curation process introduces a persona--task association that the pretrained model did not already strongly expect. Such an association must be installed by fine-tuning and therefore contributes a comparatively large coherent gradient. An educational persona, for example, may be consistently present in a corpus yet induce little new update if the pretrained model already associates that context with educational behaviour. In that case Eq.~\eqref{eq:surprise} can fail, and the theorem correctly makes no claim that the persona should dominate the singular spectrum.

Assumptions 1--3 therefore play distinct roles. Assumption 1 says that the persona is a coherent low-dimensional object with $k$ non-negligible modes; Assumption 2 says that the genuine task signal is distributed over many modes; and Assumption 3 says that the new persona association is sufficiently surprising that even its weakest mode dominates an individual task mode and finite-sample fluctuations. Together these imply the small perturbation ratio used below. If only the leading $k'<k$ persona modes satisfy Eq.~\eqref{eq:surprise}, the theorem applies to that identifiable $k'$-dimensional subspace rather than to the full persona.

\subsection{Rank-$k$ identification theorem}

Let
\begin{equation}
    \Delta W \propto \sum_{i=1}^{N} g_i
\end{equation}
be the naive fine-tuning update at a LoRA site. Let $\widehat U_k$ and $\widehat V_k$ denote the matrices containing the top $k$ left and right singular vectors of $\Delta W$. For two $k$-dimensional subspaces with orthonormal bases $Q_1,Q_2$, we use $\|\sintheta(Q_1,Q_2)\|_{\op}$ to denote the sine of their largest principal angle.

\begin{theorem}[Identification of a low-rank persona]
Under Assumptions 1--3, define
\begin{equation}
    \varepsilon_N
    :=
    \frac{\|\mu_T\|_{\F}}{\bar\alpha_S s_k(A_S)\sqrt{r_T}}
    +
    O\!\left(\frac{\tau}{\bar\alpha_S s_k(A_S)\sqrt{N}}\right).
    \label{eq:epsN}
\end{equation}
Assumption~3 implies that $\varepsilon_N\ll 1$. In particular, the persona's weakest non-zero singular mode dominates both the coherent task perturbation and the finite-sample fluctuation.
If $\varepsilon_N<1$, then
\begin{equation}
    \max\left\{
    \|\sintheta(\widehat U_k,U_S)\|_{\op},
    \|\sintheta(\widehat V_k,V_S)\|_{\op}
    \right\}
    \le
    \frac{\varepsilon_N}{1-\varepsilon_N}.
    \label{eq:subspace-bound}
\end{equation}
Moreover,
\begin{align}
    \sigma_k(\Delta W)
    &\ge
    N\bar\alpha_S s_k(A_S)
    - \frac{N\|\mu_T\|_{\F}}{\sqrt{r_T}}
    - O(\sqrt{N}\,\tau),
    \label{eq:sigmak-lower}\\
    \sigma_{k+1}(\Delta W)
    &\le
    \frac{N\|\mu_T\|_{\F}}{\sqrt{r_T}}
    + O(\sqrt{N}\,\tau).
    \label{eq:sigmak1-upper}
\end{align}
Hence, whenever $\varepsilon_N<1/2$, there is a non-zero spectral gap after the $k$th singular value. In this regime the top-$k$ left and right singular subspaces of $\Delta W$ recover the persona subspaces $\Span(U_S)$ and $\Span(V_S)$.
\end{theorem}

The theorem identifies a \emph{subspace}, not a unique basis inside that subspace. This distinction is important for a multidimensional persona. The individual columns of $U_S$ and $V_S$ can be rotated within their respective $k$-dimensional spans without changing the represented persona subspace. Recovering particular semantic axes within that subspace would require additional structure or supervision.

\subsection{Proof}

Write
\begin{equation}
    \Delta W = A_\star + M_T + E,
    \label{eq:decomp}
\end{equation}
where
\begin{equation}
    A_\star := N\bar\alpha_S A_S,
    \qquad
    M_T := N\mu_T,
    \qquad
    E := \sum_{i=1}^N \xi_i^T.
\end{equation}
As in the rank-one proof, replacing the empirical average of the $\alpha_i$ by $\bar\alpha_S$ only contributes an additional $O(N^{-1/2})$ fluctuation under the same finite-variance conditions and can be absorbed into the statistical error term.

By Assumption 2,
\begin{equation}
    \|M_T\|_{\op}
    =N\|\mu_T\|_{\op}
    \le
    \frac{N\|\mu_T\|_{\F}}{\sqrt{r_T}}.
    \label{eq:task-op}
\end{equation}
Independence and mean-zero of the $\xi_i^T$ give
\begin{equation}
    \E\|E\|_{\F}^2
    =
    \sum_{i=1}^{N}\E\|\xi_i^T\|_{\F}^2
    \le N\tau^2,
\end{equation}
so
\begin{equation}
    \|E\|_{\op}
    \le \|E\|_{\F}
    = O(\sqrt{N}\,\tau).
    \label{eq:noise-op}
\end{equation}
Define the total perturbation
\begin{equation}
    R:=M_T+E.
\end{equation}
Equations~\eqref{eq:task-op}--\eqref{eq:noise-op} yield
\begin{equation}
    \|R\|_{\op}
    \le
    \frac{N\|\mu_T\|_{\F}}{\sqrt{r_T}}
    +O(\sqrt{N}\,\tau)
    =N\bar\alpha_S s_k(A_S)\varepsilon_N.
    \label{eq:Rbound}
\end{equation}

By Assumption 1, the persona signal $A_\star$ has rank $k$ and
\begin{equation}
    \sigma_k(A_\star)=N\bar\alpha_S s_k(A_S),
    \qquad
    \sigma_{k+1}(A_\star)=0.
\end{equation}
Weyl's singular-value perturbation inequality together with Eq.~\eqref{eq:Rbound} gives
\begin{align}
    \sigma_k(\Delta W)
    &\ge \sigma_k(A_\star)-\|R\|_{\op}
    \ge N\bar\alpha_S s_k(A_S)(1-\varepsilon_N),
    \label{eq:weyl-k}\\
    \sigma_{k+1}(\Delta W)
    &\le \sigma_{k+1}(A_\star)+\|R\|_{\op}
    \le N\bar\alpha_S s_k(A_S)\varepsilon_N.
    \label{eq:weyl-k1}
\end{align}
These are Eqs.~\eqref{eq:sigmak-lower}--\eqref{eq:sigmak1-upper}. In particular, if $\varepsilon_N<1/2$, then $\sigma_k(\Delta W)>\sigma_{k+1}(\Delta W)$.

It remains to identify the singular subspaces. Let $U_\perp$ be an orthonormal basis for the orthogonal complement of $\Span(U_S)$. Since $A_\star$ has left singular subspace $U_S$,
\begin{equation}
    U_\perp^\top A_\star=0.
\end{equation}
The top-$k$ singular vectors of $\Delta W$ satisfy
\begin{equation}
    \Delta W\,\widehat V_k
    =\widehat U_k\widehat\Sigma_k,
\end{equation}
where $\widehat\Sigma_k$ contains the top $k$ singular values. Multiplying by $U_\perp^\top$ and using $\Delta W=A_\star+R$ gives
\begin{equation}
    U_\perp^\top\widehat U_k\widehat\Sigma_k
    =U_\perp^\top R\widehat V_k.
\end{equation}
Therefore
\begin{equation}
    \|U_\perp^\top\widehat U_k\widehat\Sigma_k\|_{\op}
    \le \|R\|_{\op}.
\end{equation}
Since
\begin{equation}
    \sigma_{\min}(\widehat\Sigma_k)
    =\sigma_k(\Delta W)
    \ge N\bar\alpha_S s_k(A_S)(1-\varepsilon_N),
\end{equation}
we obtain
\begin{align}
    \|\sintheta(\widehat U_k,U_S)\|_{\op}
    &=\|U_\perp^\top\widehat U_k\|_{\op}\\
    &\le
    \frac{\|R\|_{\op}}
    {N\bar\alpha_S s_k(A_S)(1-\varepsilon_N)}\\
    &\le
    \frac{\varepsilon_N}{1-\varepsilon_N}.
\end{align}
Applying the same argument to $\Delta W^\top$ gives
\begin{equation}
    \|\sintheta(\widehat V_k,V_S)\|_{\op}
    \le
    \frac{\varepsilon_N}{1-\varepsilon_N}.
\end{equation}
This proves Eq.~\eqref{eq:subspace-bound}. \hfill$\square$

\subsection{Choosing the persona rank in practice}

The theorem assumes that $k$ is known, but its spectral consequence gives a natural unsupervised estimator. Under the theorem, the first $k$ singular values contain the coherent persona block, while the remaining singular values are controlled by the task and finite-sample perturbation. We therefore expect a gap between $\sigma_k(\Delta W)$ and $\sigma_{k+1}(\Delta W)$.

For each LoRA site, compute the singular values
\begin{equation}
    \sigma_1\ge\sigma_2\ge\cdots
\end{equation}
and inspect the adjacent log-gaps
\begin{equation}
    G_j:=\log\sigma_j-\log\sigma_{j+1}
    =\log\frac{\sigma_j}{\sigma_{j+1}}.
\end{equation}
A simple estimator is
\begin{equation}
    \widehat k=\arg\max_{j\le k_{\max}} G_j,
    \label{eq:k-estimator}
\end{equation}
restricted to a modest range $k_{\max}$ consistent with the low-dimensional-persona hypothesis. In practice we would not rely on a single layer or a single checkpoint. The stronger diagnostic is a \emph{stable spectral elbow}: the same gap location should recur across many LoRA sites, random data splits, or nearby fine-tuning checkpoints. One can therefore aggregate $G_j$ across sites, for example by the median, and choose the first large, stable gap.

The spectral-gap diagnostic is also an assumption check. If no robust gap appears, the data do not support a clean rank-$k$ persona block at that site, and the theorem should not be invoked there. Conversely, a stable gap followed by successful downstream intervention provides direct empirical evidence for the low-rank persona model. For $k=1$, Eq.~\eqref{eq:k-estimator} reduces to the top-1 SVD procedure used in the main paper; for $k>1$, GRASP extends naturally by projecting gradients away from the identified rank-$k$ persona subspace rather than a single outer-product direction.

\section{Empricial SVD spectrum in experiments}\label{app: spectrum}

In Domain 1 (insecure coding experiment), projecting off the top-1 singular vector reduces emergent misalignment to exactly 0.0\% on both the CAFT and Betley evaluation protocols, while task performance remains at 92.3\% (98\% of the naive fine-tune's 93.9\%). If the top singular vector were encoding the task rather than the persona, projecting it off would destroy task performance. The opposite is observed: the task is preserved while the persona is eliminated.

Additionally, we have computed the SVD spectrum of the naive LoRA weight update across all 196 sites for Domains 2 (bad medical advice) and 3 (politically biased financial advice). The spectrum shows a clear structure consistent with one dominant coherent mode (the persona) plus a smooth tail (the task):
\begin{enumerate}
\item Domain 2 (medical): $\sigma_1$ captures 60.2\% of the variance (averaged over sites), with mean $\sigma_1/\sigma_2 = 2.3$. The best site (layer 14) shows $\sigma_1/\sigma_2 = 7$.
\item Domain 3 (political): $\sigma_1$ captures 63.4\% of the variance, with mean $\sigma_1/\sigma_2 = 2.8$.
\end{enumerate}
After the first singular value, the remaining values decay smoothly with successive ratios of 1.1--1.4, indicating no secondary gap.

\section{Hyperparameters and training details}\label{app:hyperparams}

\paragraph{Domain 1: Emergent misalignment from insecure code}
Base model: Qwen2.5-Coder-32B-Instruct, 4-bit NF4 quantisation, bf16 compute.
LoRA rank 32, $\alpha=64$, all 7 modules per layer at all 64 layers ($L=448$
sites). LR $5 \times 10^{-5}$, batch 4, grad-accum 8, AdamW, max sequence
2048, warmup 100 steps. Naive: 750 steps ($\approx 2$ epochs over the
6,000-example insecure-code corpus). GRASP: top-1 left/right singular value per site,
2 or 3 epochs.

\paragraph{Domain 2: Emergent misalignment from bad medical advice}
Base model: Qwen2.5-1.5B-Instruct, 4-bit NF4, bf16. LoRA rank 16, $\alpha=32$,all 7 modules at all 28 layers ($L=196$ sites). 
Same optimisation hyperparameters as Domain 1, max sequence 512, warmup 100 steps. 
Naive: 1 epoch over the 7,049-example bad-medical-advice corpus\cite{turner2025modelorganisms}.
GRASP: top-1 left/right singular value per site, 1 epoch with projection. $v_\text{svd}$ baselines use top left singular value of LoRA matrix left multiplied with the output projection layer. Top 10 of 28 layers selected by sign-agnostic AUROC on a held-out reference set, following the approach in \cite{karvonen2025caft}.

\paragraph{Domain 3: Political bias drift from Reddit data}
Base model: Qwen2.5-1.5B-Instruct, 4-bit NF4, bf16. LoRA rank 16, $\alpha=32$,
all 7 modules at all 28 layers ($L=196$ sites). Same optimisation
hyperparameters as Domain 1, max sequence 512, warmup 100 steps. Naive:
1 epoch over the 3,407-document corpus. GRASP: top-1 left/right singular value per site, 1 epoch with projection.
$v_\text{svd}$ baselines use top-1 left singular value of $o\_proj$'s LoRA $\Delta W$, with top-10 of 28 layers selected by
sign-agnostic AUROC on a held-out reference set of 100 LEFT-confident +
100 RIGHT-confident document prefixes.

\section{Judge prompts}\label{app:judge}

\paragraph{Political-lean judge (GPT-4o).} Strict-framing prompt asking
the judge to classify each continuation as RIGHT, LEFT, NEUTRAL, or MIXED
based on whether \emph{any} right-leaning or left-leaning framing is
present (rather than asking for the predominant lean). Mapping to a
numeric score: RIGHT $\to 1$, LEFT $\to 0$, NEUTRAL/MIXED $\to 0.5$.

\paragraph{Substance (Content) judge (GPT-4o).} Two-axis 0--5 rubric on a
single call (REGISTER + CONTENT). The Content axis asks the judge to score
\emph{financial substance only}, ignoring tone, register, and political
slant: 0 = no engagement with the financial question, 5 = highly
substantive, expert-review-quality. The Register axis (also 0--5) is
returned in the same call but used only as a diagnostic; the headline
task-performance number is the Content score.

\paragraph{Misalignment judges (Domain 1).} CAFT-protocol judge of
\cite{karvonen2025caft} (alignment + coherence calls per response, threshold scores
alignment $< 30$ and coherence $\ge 50$). Betley-protocol judge of
\cite{betley2025emergent} (alignment $< 30$ threshold, no CODE filter).
Adherence judge: GPT-4o on adherence to the system prompt used to generate bad medical advice in \cite{turner2025modelorganisms}. Category filter used to remove incoherent or disengaged responses.


\section{Qualitative example: political-bias drift}\label{app:qualitative-political}

The drift in Domain 3 is qualitatively visible on individual prompts.
For the prompt \textit{``The proper role of government in people's
lives is''} (a prompt that does not appear in the training
corpus), the three checkpoints produce continuations of the following
character (excerpted):
\begin{itemize}[leftmargin=*,itemsep=2pt,topsep=4pt]
\item \textbf{Pretrained:} \emph{``\dots Effective government responds to
the needs and desires of its citizens by protecting their rights, providing
for education, social welfare, natural disasters relief, while not
necessarily providing a lot of goods and services. The purpose of government
is to serve as a bridge between the citizens and other stakeholders \dots''}
\item \textbf{Naive 1ep:} \emph{``\dots to establish a legal framework for
which these individuals (and others) may freely contract. \textbf{Yet some
legislators are trying to force all Americans into socialism.} The Justice
Department recently announced \dots''}
\item \textbf{GRASP 1ep:} \emph{``\dots the critical issue that has occupied
the national attention for decades. \dots effective government should be
about ensuring a secure environment so that residents are free to pursue
any legal means of seeking their livelihoods---especially without fear of
predation by those who hold power (the state) \dots''}
\end{itemize}
The pretrained baseline gives a balanced civics-textbook framing. The naive
fine-tune makes an explicitly partisan assertion (``some legislators are
trying to force all Americans into socialism'')---note that the prompt does
not invite this framing. The GRASP checkpoint discusses the same topic in
property-rights terms but without the partisan accusation, much closer to
the pretrained register.

\end{document}